%% file: IEEETMI_HARDIReg.tex
\documentclass[12pt,journal,draftcls,letterpaper,onecolumn]{IEEEtran}
\usepackage{times}
\usepackage{epsfig}
\usepackage{graphicx}
\usepackage{amsmath}
\usepackage{amssymb}
\usepackage{amsthm}
\usepackage{makeidx}  
\usepackage{url}
\usepackage{algorithm}
\usepackage{bm}
\usepackage{microtype}
\usepackage{graphicx}
\usepackage{url}
\usepackage[footnotesize]{caption}
\usepackage[footnotesize]{subfigure}
\usepackage[rightcaption]{sidecap}
\usepackage{pdflscape}
\usepackage[pdftex,bookmarks=true]{hyperref}
\usepackage{lineno}
\usepackage{ulem}
\usepackage{afterpage}
\usepackage[usenames,dvipsnames]{color}

\newtheorem{theorem}{Theorem}[section]
\newtheorem{lemma}{Lemma}[section]
\input{HARDI_NIMG2011_macro}
\begin{document}

\title{Diffeomorphic Metric Mapping of High Angular Resolution Diffusion Imaging based on Riemannian Structure of Orientation Distribution Functions}
%
%
%

\author{Jia~Du,~
        Alvina~Goh,~\IEEEmembership{Member,~IEEE,}
        and~Anqi~Qiu~ 
\thanks{J. Du is with the Division of Bioengineering, National University of Singapore, Singapore.}
\thanks{A. Goh is with the Department of Mathematics,, National University of Singapore, Singapore.}
\thanks{A. Qiu is with the Division of Bioengineering and Clinical Imaging Research Center, National University of Singapore, Singapore and the Singapore Institute for Clinical Sciences, Agency for Science, Technology and Research e-mail: bieqa@nus.edu.sg.}
\thanks{Manuscript received June 2011.}}

\markboth{IEEE Transactions on Medical Imaging,~Vol.~X, No.~X, XXX~2011}%
{IEEE Transactions on Medical Imaging,~Vol.~X, No.~X, XXX~2011}%

\maketitle

\begin{abstract}
In this paper, we propose a novel large deformation diffeomorphic registration algorithm to align high angular resolution diffusion images (HARDI) characterized by orientation distribution functions (ODFs). Our proposed algorithm seeks an optimal diffeomorphism of large deformation between two ODF fields in a spatial volume domain and at the same time, locally reorients an ODF in a manner such that it remains consistent with the surrounding anatomical structure. To this end, we first review the Riemannian manifold of ODFs. We then define the reorientation of an ODF when an affine transformation is applied and subsequently, define the diffeomorphic group action to be applied on the ODF based on this reorientation.  We incorporate the Riemannian metric of ODFs for quantifying the similarity of two HARDI images into a variational problem defined under the large deformation diffeomorphic metric mapping (LDDMM) framework. We finally derive the gradient of the cost function in both Riemannian spaces of diffeomorphisms and the ODFs, and present its numerical implementation. Both synthetic and real brain HARDI data are used to illustrate the performance of our registration algorithm.

\end{abstract}


%
\IEEEpeerreviewmaketitle

\section{Introduction}
\label{sec:intro}

Diffusion weighted magnetic resonance imaging (DW-MRI) is a unique \emph{in vivo} imaging technique that allows us to visualize the three-dimensional architecture of neural fiber pathways in the human brain. Several techniques may be used to reconstruct the local orientation of brain tissue from DW-MRI data. A classical method is known as Diffusion Tensor Imaging (DTI) \cite{Basser:JMR94}, which characterizes the diffusivity profile of water molecules in brain tissue by a single oriented 3D Gaussian probability distribution function (PDF). In DTI, the diffusivity profile is often represented mathematically by a symmetric positive definite (SPD) tensor field $\D:\Re^3 \to \SPD(3) \subset \Re^{3\times 3}$ that measures the extent of diffusion in any direction $\v\in\Re^3$ as $\v^\top \D\v$. The geometry of $\SPD(3)$ is well-studied and several metrics for comparing tensors have been proposed \cite{Pennec:MRM06,Kindlmann:MICCAI07,Lenglet:JMIV06,Pennec:IJCV06}. Based on these metrics, statistical tests such as voxel-based analysis of 
diffusion tensors have been developed \cite{Buchel:Cereb04,Wang:AJNeuro06,Hua:neuro08,Jahanshad:neuro10}. Before such population studies can been carried out, there is a essential need to perform DTI registration, that is, to align tensor data across subjects to a standard coordinate space.

Compared to the classical image registration problem, the registration of DTI fields is more complicated since DTI data contains structural information affected by the transformation. Two key transformations need to be defined: a transformation to spatially align anatomical structures between two brains in a $3$D volume domain, and a transformation to align the local diffusivity profiles defined at each voxel of two brains. More precisely, a transformation $\phi$ of the image domain induces a reorientation of the DTI as the direction of diffusion depends on the coordinate system. Thus, for two diffusion tensors $\D_1(x)$ and $\D_2(x)$ at voxel $x$, it is no longer true that $\D_1(x) \approx \D_2(\phi(x))$ and each tensor must be reoriented in such a way that it remains consistent with the surrounding anatomical structure. There exist several approaches for reorientation that are used in DTI \cite{Alexander:TMI01}. For instance, the Finite Strain (FS) scheme decomposes an affine transformation matrix $A$ into $A=RS$, where $R$ is the rigid rotation and $S$ is the deformation, and reorients the tensor $\D$ as $R\D R^\top$. An alternative strategy is the Preservation of Principal Direction (PPD), in which the reoriented tensor keeps its eigenvalues, yet its principal eigenvector $\v_1$ is transformed as $A\v_1/\|A\v_1\|$. The reader is referred to \cite{Cao:TMI06,Goh:ECCV06,Guimond:ISBI02,Ruiz:MEDIA02,Zhang:MIA06,Chiang:TMI08} and references therein for the existing DTI registration methods. 


While it has been demonstrated that DTI is valuable for studying brain white matter development in children and detecting abnormalities in patients with neuropsychiatric disorders and neurodegenerative diseases, a major shortcoming of DTI is that it can only reveal one dominant fiber orientation at each location, when between one and two thirds of the voxels in the human brain white matter are thought to contain multiple fiber bundles crossing each other \cite{Behrens:NeuroImage07}. High angular resolution diffusion imaging (HARDI) \cite{Tuch:MRM02b} addresses this well-known limitation of DTI. HARDI measures diffusion along $n$ uniformly distributed directions on the sphere and can characterize more complex fiber geometries. Several reconstruction techniques can be used to characterize diffusion based on the HARDI signals. One class is based on higher-order tensors \cite{Barmpoutis:Neuroimage09,Ghosh:MICCAI08} and leverage prior work on DTI. Another method is Q-ball Imaging, which uses the Funk-Radon transform to reconstruct an orientation distribution function (ODF). The model-free ODF is the angular profile of the diffusion PDF of water molecules and has been approximated using different sets of basis functions such as spherical harmonics (SH) \cite{Ozarslan:MRM03,Descoteaux:MRM07,Frank:MRM02,Hess:MRM06,Aganj:MRM10}. Such methods are relatively fast to implement because the ODF is computed analytically. By quantitatively comparing fiber orientations retrieved from ODFs against histological measurements, Leergaard et al. \cite{leergaard_plosone_2010} shows that accurate fiber estimates can be obtained from HARDI data, further validating its usage in brain studies.

Similar to the case of DTI, an open challenge in the analysis of mathematically complex HARDI data is registration. Several HARDI registration algorithms have been recently proposed under a specific model of local diffusivity.  
Chiang et al. \cite{Chiang:MICCAI08} proposes an information-theoretic approach for fluid registration of ODFs. An inverse-consistent fluid registration algorithm that 
minimizes the symmetrized Kullback-Leibler divergence (sKL) or J-divergence of the two DT images \cite{Chiang:TMI08} is first performed and the ODF fields are registered by applying the corresponding DTI mapping. The ODFs are reoriented using the PPD method where the principal direction of the ODF is determined by principal component analysis.
Cheng et al. \cite{ChengG:MICCAI09} takes the approach of representing HARDI by Gaussian mixture fields (GMF) and assumes a thin-plate spline deformation. The $\bL^2$ metric of GMFs is minimized, and reorientation is performed on the individual Gaussian components, each representing a major fiber direction. 
Barmpoutis et al. \cite{Barmpoutis:MICCAI07} uses a $4$th order tensor model and assumes a region-based nonrigid deformation. The rotationally invariant Hellinger distance is considered and an affine tensor reorientation, which accounts for rotation, scaling and shearing effects, is applied.  Geng et al. \cite{Geng:TMI11} performs a diffeomorphic registration is performed with the $\bL^2$ metric on ODFs represented by spherical harmonics. Reorientation is done by altering the SH coefficients in a manner similar to the FS method in DTI where only the rotation is extracted and applied. Bloy et al. \cite{Bloy:ISBI10} performs alignment of ODF fields by using a multi-channel diffeomorphic demons registration algorithm on rotationally invariant feature maps and uses the FS scheme in reorientation.  Yap et al. \cite{Yap:Neuro2011} uses the SH-based ODF representation and proposes a hierarchical registration scheme, where descriptors are extracted at each level and the alignment is updated by using features extracted from the increasing order of the SH representation. Reorientation is done by tilting the gradient directions via multiplying with the local affine transform and normalizing.

\bigskip\noindent\textbf{Paper Contributions.} Unlike a majority of the above-mentioned HARDI registration approaches that seek small deformation between two brains, we present a novel registration algorithm for HARDI data represented by ODFs under the framework of large deformation diffeomorphic metric mapping (LDDMM) such that the deformation of two brains is diffeomorphic (one-to-one, smooth, and invertible) and can be in a large scale. Previous studies \cite{miller_pnas_2005} suggest that the transformation from one brain to another can be really large and therefore small deformation models may not be enough. Our proposed algorithm seeks an optimal diffeomorphism of large deformation between two ODF fields across a spatial volume domain and at the same time, locally reorients an ODF in a manner that it remains consistent with the surrounding anatomical structure. We define the reorientation of an ODF when an affine transformation is applied and subsequently, define the diffeomorphic group action to be applied on the ODF based on this reorientation. The ODF reorientation used in this paper ensures that the transformed ODF remains consistent with the surrounding anatomical structure and at the same time, not solely dependent on the rotation. Rather, the reorientation takes into account the effects of the affine transformation and ensures the volume fraction of fibers oriented toward a small patch must remain the same after the patch is transformed. The Riemannian metric for the similarity of ODFs is then incorporated into a variational problem in LDDMM. Finally, we derive the gradient of the cost function in both Riemannian spaces of diffeomorphisms and the ODFs and present its numerical implementation. Even though this paper is based on our previous work \cite{Du:HARDI}, one major fundamental difference is that the gradient derivation in this paper account for orientation differences in the ODFs while \cite{Du:HARDI} does not. We will elaborate how the proposed algorithm outperforms that in \cite{Du:HARDI} while we discuss the gradient derivation in in \S \ref{sec:gradODF}. Our experiments are shown on synthetic and real HARDI brain data in \S \ref{sec:expts}.

\section{Methods}
\subsection{Review: the Riemannian Manifold of ODFs}
\label{sec:RieODF}

As mentioned in \S \ref{sec:intro}, HARDI measurements can be used to reconstruct the ODF, the angular profile of the diffusion probability density function (PDF) of water 
molecules. The ODF is actually a PDF defined on a unit sphere $\Do^2$ and its space is defined as
\begin{align*}
\P=
\{\p:\Do^2\rightarrow \Re^+ | \forall \s\in\Do^2, \p(\s) \geq 0; \int_{\s\in\Do^2} \p(\s) d\s=1\}  \ .
\end{align*}
The space of $\p$ forms a Riemannian manifold, also known as the statistical manifold, which is well-known from the field of \emph{information geometry} \cite{Amari85}. Rao \cite{Rao:BCMS45} introduced the notion of the statistical manifold whose elements are probability density functions and composed the Riemannian structure with the \emph{Fisher-Rao} metric. \cite{Cencov82} showed that the Fisher-Rao metric is the \emph{unique intrinsic metric} on the statistical manifold $\P$ and therefore invariant to re-parameterizations of the functions. There are many different parameterizations of PDFs that are equivalent but with different forms of the Fisher-Rao metric, leading to the Riemannian operations having different computational complexity. In this paper, we choose the square-root representation, which is used recently in ODF processing \cite{Goh:NeuroImage2011,Cheng:MICCAI09}. The square-root representation is one of the most efficient representations found to date as the various Riemannian operations such as geodesics, exponential maps, logarithm maps are available in closed form. 

The \emph{square-root ODF} ($\gsODF$) is defined as
$\displaystyle
\bpsi(\s)=\sqrt{\p(\s)}$, where $\bpsi(\s)$ is assumed to be non-negative to ensure uniqueness. The space of such functions is defined as
\begin{align}
\label{eq:sphereeqn}
\bPsi=
\{\bpsi:\Do^2\rightarrow \Re^+ | \forall \s\in\Do^2, \bpsi(\s) \geq 0; \int_{\s\in\Do^2} \bpsi^2(\s) d\s=1\}.
\end{align}
We see that from Eq. \eqref{eq:sphereeqn}, the functions $\bpsi$ lie on the positive orthant of a unit Hilbert sphere, a well-studied Riemannian manifold. It can be shown \cite{Srivastava:CVPR07} that the Fisher-Rao metric is simply the $\bL^2$ metric, given as
\begin{align}
\label{eq:FisherRao}
\langle \bxi_j, \bxi_k \rangle_{\bpsi_i} = 
\int_{\s\in\Do^2} \bxi_j(\s) \bxi_k (\s) d\s,
\end{align}
where $\bxi_j, \bxi_k \in T_{\bpsi_i} \bPsi$ are tangent vectors at $\bpsi_i$. The geodesic distance between  any two functions $\bpsi_i, \bpsi_j \in \bPsi$ on a unit Hilbert sphere is the angle
\begin{align}
\dist(\bpsi_i,\bpsi_j)=
\|\log_{\bpsi_i}(\bpsi_j)\|_{\bpsi_i} 
=\cos^{-1} \langle \bpsi_i, \bpsi_j \rangle = 
\cos^{-1}\left(\int_{\s\in\Do^2} \bpsi_i(\s) \bpsi_j(\s)d\s\right),
\end{align}
where $\langle \cdot, \cdot \rangle$ is the normal dot product between points in the sphere under the $\bL^2$ metric. For the sphere, the \emph{exponential map} has the closed-form formula
\begin{align}
\label{eq:exp}
\exp_{\bpsi_i}(\bxi) = \cos (\|\bxi\|_{\bpsi_i}) \bpsi_i + \sin(\|\bxi\|_{\bpsi_i})\frac{\bxi}{\|\bxi\|_{\bpsi_i}},
\end{align}
where $\bxi \in T_{\bpsi_i}\bPsi$ is a tangent vector at $\bpsi_i$ and 
$\|\bxi\|_{\bpsi_i}=\sqrt{\langle \bxi,\bxi\rangle_{\bpsi_i}}$. By restricting $\|\bxi\|_{\bpsi_i}\in [0,\frac{\pi}{2}]$, we ensure that the exponential map is bijective. The \emph{logarithm map} from $\bpsi_i$ to $\bpsi_j$ has the closed-form formula
\begin{align}
\label{eq:log}
\overrightarrow{\bpsi_i \bpsi_j}&=\log_{\bpsi_i}(\bpsi_j)
=\frac{\bpsi_j - \langle \bpsi_i,\bpsi_j \rangle \bpsi_i}{\sqrt{1-\langle \bpsi_i,\bpsi_j\rangle^2}} \cos^{-1} \langle \bpsi_i,\bpsi_j \rangle.
\end{align}

\subsection{Affine Transformation on Square-Root ODFs}
\label{sec:LocalAffine}

In this section, we discuss the reorientation of the $\gsODF$, $\bpsi(\s)$, when an affine transformation $A$ is applied. We denote the transformed $\gsODF$ as $\widehat\bpsi(\widehat{\s}) = A \bpsi(\s)$, reflecting the fact that an affine transformation induces changes in both the magnitude of $\bpsi$ and the gradient directions of $\s$. We will now illustrate how the reorientation is done.

\begin{figure}[htb]
\centering
\includegraphics[width=0.6\textwidth]{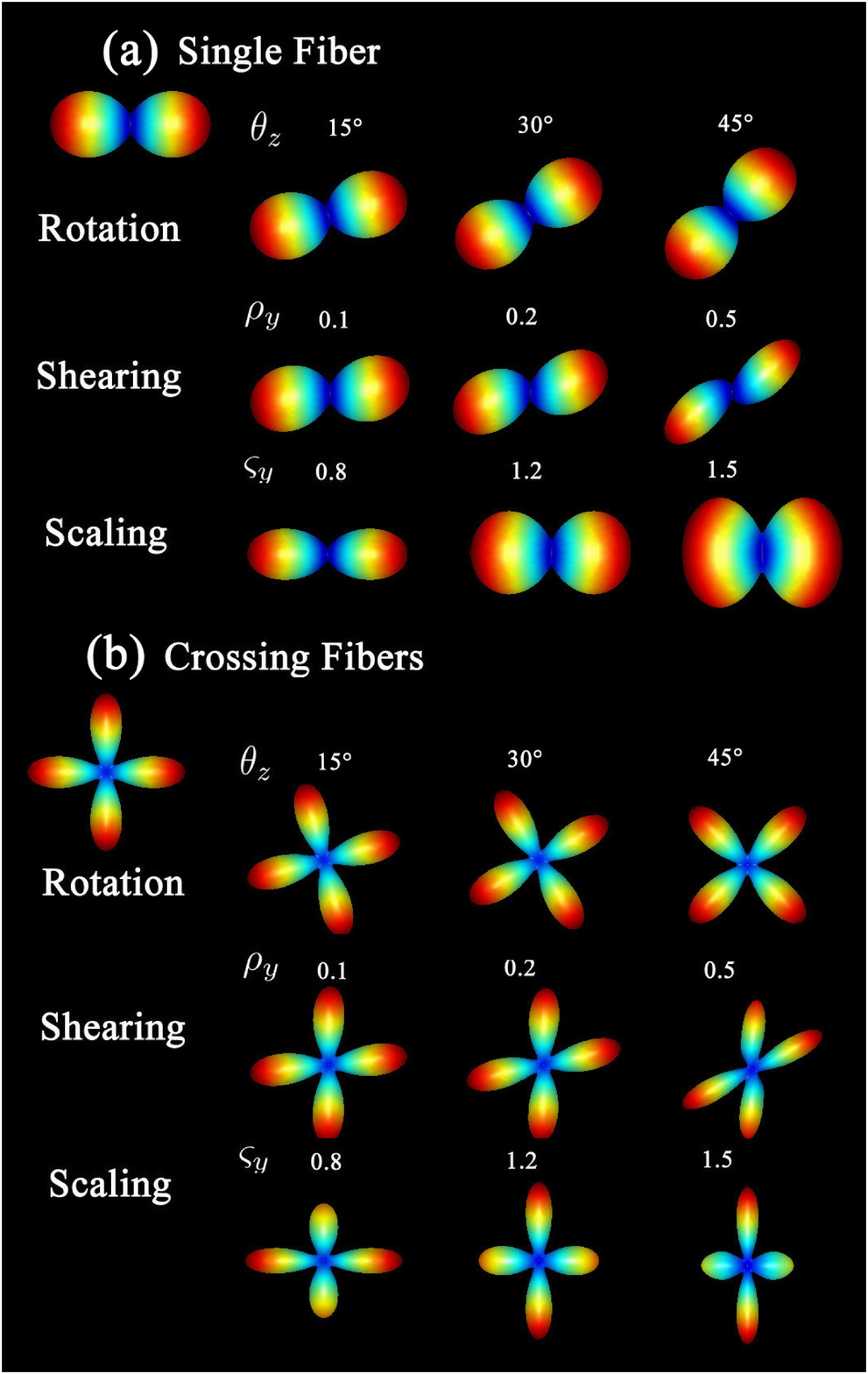}
\caption{Examples of local affine transformations on ODFs with a single orientation fiber (panel (a)) and crossing fibers (panel (b)). From top to bottom of each panel, three types of affine transformations, $A$, on the ODFs are demonstrated: a rotation with angle $\theta_z$, where $A=[\cos\theta_z \: -\sin\theta_z \: 0 ;\;  \sin\theta_z \: \cos\theta_z \: 0 ;\:  0 \: 0 \: 1]$; a vertical shearing with factor $\rho_y$, where $A=[1 \; 0 \; 0;\;  -\rho_y \; 1 \; 0 ;\;  0 \; 0 \; 1]$; and a vertical scaling with factor $\varsigma_y$ where $A=[1 \; 0 \; 0 ;\;  0 \; \varsigma_y \; 0 ;\;  0 \; 0 \; 1 ]\ .$}
\label{Fig:AffineAll}
\end{figure}

First of all, we discuss the change in the gradient directions of $\s$. We assume that the change of the gradient directions due to affine transformation $A$ is
\begin{align}
\label{eq:transcoord}
 \widehat{\s}&=\frac{A^{-1}\s}{\|A^{-1}\s\|} \ ,
\end{align}
where the transformed gradient directions $\widehat{\s}$ are normalized back into the unit sphere $\Do^2$. Notice that for $\s\in\Do^2$, Eq. \eqref{eq:transcoord} defines an invertible function of $\s$ and therefore, we can find the ODF $A \bpsi(\s)$ using the change-of-variable technique of PDF. This will give us the following theorem. 
\begin{theorem}
\textbf{Reorientation of $\bpsi$ based on affine transformation of $A$.}
\label{thm:reorient}
 Let $A\bpsi(\s)$ be the result of an affine transformation $A$ acting on a $\gsODF$ $\bpsi(\s)$. The following analytical equation holds true%
\begin{align}
\label{eq:theorem}
A\bpsi(\s)=\sqrt{\frac{\det{A^{-1}}}{\left\|A^{-1}\s\right\|^3}} \quad \bpsi\left(\frac{A^{-1}\s}{\|A^{-1}\s\|}\right),
\end{align}
where $\left\|\cdot\right\|$ is the norm of a vector. 
\end{theorem}
The ODF reorientation used in this paper ensures that the transformed ODF remains consistent with the surrounding anatomical structure and at the same time, not solely dependent on the rotation. Rather, by constructing the change-of-variable technique, the reorientation takes into account the effects of the affine transformation and ensures the volume fraction of fibers oriented toward a small patch must remain the same after the patch is transformed. 
Figure \ref{Fig:AffineAll} illustrates how $A\bpsi(\s)$ varies when $A$ is a rotation, shearing, or scaling and $\bpsi(\s)$ contains a single fiber or crossing fibers. By construction, $A\bpsi(\s)$ fulfills the definition of the $\gsODF$. Hence, the similarity of $A\bpsi(\s)$ to the square-root ODFs can be quantified in the Riemannian structure given in \S \ref{sec:RieODF} for the HARDI registration. 

\afterpage{\clearpage}

\subsection{Diffeomorphic Group Action on Square-Root ODF}
\label{sec:DiffGrpAct}

We have shown in \S \ref{sec:LocalAffine} how to reorient $\bpsi$ located at a fixed spatial position $x$ in the image volume $\bOmega\subset\Re^3$ through an affine transformation. In this section, we define an action of diffeomorphisms $\phi : \bOmega \rightarrow \bOmega$ on  $\bpsi$, which takes into consideration the reorientation of $\bpsi$ as well as the transformation of the spatial volume in $\bOmega \subset \Re^3$.  Denote $\bpsi(\s,x)$ as the $\gsODF$ with the orientation direction $\s\in\Do^2$ located at $x\in\bOmega$. We define the action of diffeomorphisms on $\bpsi(\s,x)$  in the form of 
\begin{align*}
\phi\cdot\bpsi(\s,x)  =  A_{\phi^{-1}(x)}\bpsi(\s,\phi^{-1}(x)),
\end{align*}
where the local affine transformation $A_{x}$ at spatial coordinates $x$ is defined as the Jacobian matrix of $\phi$ evaluated at $x$, \ie $A_{x}= D_{x}\phi$. According to Eq. \eqref{eq:theorem}, the action of diffeomorphisms on $\bpsi(\s,x)$ can be computed as
\begin{align}
\label{eq:groupaction}
\phi\cdot\bpsi(\s,x)  =   \sqrt{\frac{\det{\bigl(D_{\phi^{-1}}\phi \bigr)^{-1}} }{\left\|{\bigl(D_{\phi^{-1}}\phi \bigr)^{-1} } \s \right\|^3} } \quad
\bpsi \left( \frac{(D_{\phi^{-1}}\phi \bigr)^{-1} \s}{\|(D_{\phi^{-1}}\phi \bigr)^{-1} \s\|}, \phi^{-1}(x) \right) .
\end{align}
For the sake of simplicity,  we denote $\phi\cdot\psi(\s,x)$  as
\begin{align}
\label{eq:subtran}
\phi\cdot\bpsi(\s,x) =  A\bpsi\circ\phi^{-1}(x) \ ,
\end{align}
where it will be used in the rest of the paper.

Since $\phi\cdot\bpsi(\s,x)$ is in the space of $\gsODF$,  the Riemannian distance given in \S \ref{sec:RieODF} can be directly used to quantify the similarity of $\phi\cdot\bpsi(\s,x)$ to other $\gsODF$s, which we employ in the HARDI registration described in the following section.

\subsection{Large Deformation Diffeomorphic Metric Mapping for ODFs}
\label{sec:LDDMMODF}

The previous sections equip us with an appropriate representation of the ODF and its diffeomorphic action. Now, we state a variational problem for mapping ODFs from one volume to another. We define this problem in the ``large deformation" setting of Grenander's group action approach for modeling shapes, that is, ODF volumes are modeled by assuming that they can be generated from one to another via flows of diffeomorphisms $\phi_t$, which are solutions of ordinary differential equations 
$\dot \phi_t = v_t (\phi_t), t \in [0,1],$ starting from
the identity map $\phi_0={\Id}$. They are therefore characterized by
time-dependent velocity vector fields $v_t, t \in [0,1]$. We define a metric distance between a target volume $\bpsi_{\targ}$ and a template volume $\bpsi_{\temp}$ as the minimal length of curves $\phi_t \cdot \bpsi_{\temp}, t \in [0,1],$ in a shape space such that, at time $t=1$,  $\phi_1 \cdot
\bpsi_{\temp} = \bpsi_{\targ}$. Lengths of such curves are computed as the integrated norm $\| v_t \|_V$ of the
vector field generating the transformation, where $v_t \in V$, where $V$ is a
reproducing kernel Hilbert space with kernel $k_V$ and norm $\| \cdot \|_V$.

To ensure solutions are diffeomorphisms, $V$ must be a space of smooth
vector fields \cite{DuGrMi1998}.
Using the duality isometry in Hilbert spaces, one can equivalently express the lengths in terms of $m_t$, interpreted as momentum such that for each $u\in
V$, 
\begin{align}
\langle m_t, u \circ \phi_t\rangle_2 = \langle k_V^{-1}v_t, u\rangle_2,
\end{align}
where we let $\langle m, u\rangle_2$ denote the $\bL^2$ inner product between $m$ and
$u$, but also, with a slight abuse, the result of the natural pairing
between $m$ and $v$ in cases where $m$ is singular (e.g., a
measure). This identity is classically written as $\phi_t^* m_t =
k_V^{-1} v_t$, where $\phi_t^*$ is referred to as the pullback
operation on a vector measure, $m_t$. 
Using the identity $\|v_t\|_V^2 = \langle k_V^{-1}v_t,
v_t\rangle_2=\langle m_t,k_Vm_t\rangle_2$ and the standard fact that energy-minimizing curves
coincide with constant-speed length-minimizing curves, one can obtain 
the metric distance between the template and target $\gsODF$ volumes, $ \rho(\bpsi_{\temp},\bpsi_{\targ})$, by minimizing 
$\int_0^1 \langle m_t, k_Vm_t\rangle_2 dt$ such that $\phi_1 \cdot \bpsi_{\temp}=\bpsi_{\targ} $ at time $t=1$. 

We associate this with the variational problem in the form of  
\begin{align}
\label{eqn:metric} 
J(m_t) = \inf_{\begin{subarray}{l}m_t: \dot \phi_t =
k_Vm_t(\phi_t),\\ \quad\phi_0=\Id\end{subarray}} \rho(\bpsi_{\temp},\bpsi_{\targ})^2 
+ \lambda \int_{x \in \bOmega} E_x(\phi_1 \cdot \bpsi_{\temp}(\s,x),\bpsi_{\targ}(\s,x)) dx
\end{align}
with $E_x$ as the metric distance between 
the deformed $\gsODF$ template, $\phi_1 \cdot \bpsi_{\temp}(\s,x)$, and the target, $\bpsi_{\targ}(\s, x)$. We use the Riemannian metric given in \S \ref{sec:RieODF} and rewrite Eq. \eqref{eqn:metric} as 
\begin{align}
\label{eqn:lddmm} 
J(m_t) = \inf_{\begin{subarray}{l}m_t: \dot \phi_t =
k_Vm_t(\phi_t),\\ \quad\phi_0=\Id\end{subarray}}\int_0^1 \langle m_t, k_Vm_t\rangle_2 dt 
 +\lambda \int_{x\in\bOmega}\|\log_{A\bpsi_{\temp} \circ \phi_1^{-1}(x)}(\bpsi_{\targ}(x))\|^2_{A\bpsi_{\temp}\circ \phi_1^{-1}(x)}dx, 
\end{align}
where $A=D\phi_1$, the Jacobian of $\phi_1$. For the sake of simplicity, we denote $\bpsi_{\targ}(\s,x)$ as 
$\bpsi_{\targ}(x)$. Note that since we are dealing with vector fields in $\mathbb{R}^3$,
the kernel of $V$ is a matrix kernel operator in order to get a proper definition. We define this kernel as $k_V \Id_{3 \times 3}$, where $\Id_{3 \times 3}$ is an identity matrix, such that $k_V$ can be a scalar kernel. In the rest of the paper, we shall refer to this LDDMM mapping problem as LDDMM-ODF.

\subsection{Gradient of $J$ with respect to $m_t$}
\label{sec:gradODF}
The gradient of $J$ with respect to $m_t$ can be computed via studying a variation ${m}_t^\epsilon=m_t + \epsilon \widetilde{m}_t $ on $J$ such that the derivative of $J$ with respect to $\epsilon$ is expressed in function of $ \widetilde{m}_t$. According to the general LDDMM framework derived in \cite{Du:SD,anqi_joan_2007},  we directly give the expression of {\bf the gradient of $J$ with respect to $\boldsymbol{m_t}$} as 
\begin{eqnarray}
\label{eqn:gradJ}
\nabla J(m_t) &= & 
 2 m_t + \lambda \eta_t \ ,
\end{eqnarray}
where 
\begin{equation}
\label{eqn:eta}
\eta_t = \nabla_{\phi_1}E +  \int_t^1 \bigl[ \partial_{\phi_s} (k_V m_s) \bigr]^\top (\eta_s + m_s) ds \ ,
\end{equation}
where $\partial_{\phi_s} (k_V m_s)$ is the partial derivative of $k_Vm_s$ with respect to $\phi_s$. 
$\eta_t$ in Eq. \eqref{eqn:eta} can be solved backward given $\eta_1= \nabla_{\phi_1}E$, where $E=\int_{x\in \bOmega}E_x dx$, which will be discussed in the following.

\bigskip\noindent {\bf Gradient of $E$ with respect to $\phi_1$: } 
The computation of $\nabla_{\phi_1} E$ is not straightforward and the Riemannian structure of ODFs has to be incorporated. Let's first compute $\nabla_{\phi_1} E_x$ at a fixed location, $x$. We consider a variation $\phi_1^\epsilon=\phi_1 + \epsilon h $ of $\phi_1$ and denote the corresponding variation in $A$ as $A^\epsilon$, where $A=D_x\phi_1$ and $A^\epsilon=D_x\phi_1^\epsilon \ .$
Here, we directly give the expression of $\partial_\epsilon E_x |_{\epsilon=0}$ and the reader is referred to Appendix  \ref{app:gradient} for the full derivation of terms (A) and (B) in the following equation.

\begin{align}
\label{eq:gradEx}
& \partial_\epsilon E_x |_{\epsilon=0}  \\ 
\nonumber
 = & 2\Bigl\langle \log_{A\bpsi_{\temp} \circ \phi_1^{-1}(x)} \bpsi_{\targ}(x), \frac{\partial \log_{A^\epsilon\bpsi_{\temp} \circ (\phi_1^\epsilon)^{-1}(x)} \bpsi_{\targ}(x)}{\partial \epsilon} |_{\epsilon=0} \Bigr\rangle_{A\bpsi_{\temp}\circ \phi_1^{-1}(x) }\\
\nonumber
 = & -2\Bigl\langle \log_{A\bpsi_{\temp} \circ \phi_1^{-1}(x)} \bpsi_{\targ}(x),\frac{\partial\log_{A\bpsi_{\temp} \circ \phi_1^{-1}(x)}A^\epsilon \bpsi_{\temp} \circ (\phi_1^\epsilon)^{-1}(x)}{\partial \epsilon} |_{\epsilon=0} \Bigr\rangle_{A\bpsi_{\temp}\circ \phi_1^{-1}(x) }\\  \nonumber
 = & \underbrace{-2\Bigl\langle \log_{A\bpsi_{\temp} \circ \phi_1^{-1}(x)} \bpsi_{\targ}(x), \frac{\partial \log_{A\bpsi_{\temp} \circ \phi_1^{-1}(x)}A \bpsi_{\temp} \circ (\phi_1^\epsilon)^{-1}(x)}{\partial \epsilon}|_{\epsilon=0}  \Bigr\rangle_{A\bpsi_{\temp}\circ \phi_1^{-1}(x) } }_{\text{term \ (A)}}\\  \nonumber
 &  \underbrace{-2\Bigl\langle \log_{A\bpsi_{\temp} \circ \phi_1^{-1}(x)} \bpsi_{\targ}(x), 
\frac{\partial \log_{A\bpsi_{\temp} \circ \phi_1^{-1}(x)}A^\epsilon \bpsi_{\temp} \circ (\phi_1)^{-1}(x)}{\partial \epsilon}|_{\epsilon=0}  \Bigr\rangle_{A\bpsi_{\temp}\circ \phi_1^{-1}(x) }}_{\text{term \ (B)}}  
\end{align}

\begin{align*}
\nonumber
 =& 2 \det{\phi_1(x)}\Biggl\{\underbrace{\Bigl\langle (D_x\phi_1)^{-\top} \bigl\langle \log_{A\bpsi_{\temp}(x)} \bpsi_{\targ}(\phi_1(x)), \nabla_{x}(A\bpsi_{\temp})\bigr\rangle_{ A\bpsi_{\temp}(x)}, h\Bigr\rangle}_{\text{term \ (A)}} \\ \nonumber
 &+\underbrace{\sum^3_{i=1}\Bigl\langle \divo\bigl(\bigl\langle \log_{A\bpsi_{\temp}(x)} \bpsi_{\targ}(\phi_1(x)), L_x^i \bigr\rangle_{ A\bpsi_{\temp}(x)}\bigr) \e^i,h \Bigr\rangle}_{\text{term \ (B)}} \Biggr\} \ ,
\end{align*}
where $^\top$ denotes the matrix transpose and  $\e^i$ is a $3 \times 1$ vector with the $i$th element as one and the rest as zero. 
$ \langle\cdot,\cdot\rangle_{A\bpsi_{\temp}(x)}$ is the Fisher-Rao metric defined in Eq. \eqref{eq:FisherRao}. $\nabla_{x}(A\bpsi_{\temp})$ in term (A) is the first derivative of the  $\gsODF$, $A\bpsi_{\temp}$, with respect to $x$. Since $A\bpsi_{\temp}$ also lies in the Riemannian manifold of $\gsODF$s, $\nabla_x (A\bpsi_{\temp})$ is a vector with each element being a logarithm map of $A\bpsi_{\temp}$ and is defined as
\begin{align*}
\nabla_{x}\left[A\bpsi_{\temp}(x)\right]&=\left[\begin{array}{c} \frac{1}{|\triangle e_1|}\log_{A\bpsi_{\temp}(x)}A\bpsi_{\temp}(x+\triangle e_1) \\ \frac{1}{|\triangle e_2|}\log_{A\bpsi_{\temp}(x)}A\bpsi_{\temp}(x+\triangle e_2) \\ \frac{1}{|\triangle e_3|}\log_{A\bpsi_{\temp}(x)}A\bpsi_{\temp}(x+\triangle e_3) \end{array}\right],
\end{align*}
where $\triangle e_1,\triangle e_2$ and $\triangle e_3$ indicate small variations in three orthonormal directions of $\Re^3$, respectively. 

In term (B) of Eq. \eqref{eq:gradEx}, we define $L_x$ as a $3 \times 3$ matrix of logarithm maps with its $i$th column written as 
$$
L_x^i=(D_x\phi_1)^{-1}\s u^i-\frac{1}{2}A\bpsi_{\temp}(\s,x)\w^i \ ,
$$
where $\w^i$ is the $i$th column of $(D_x \phi_1)^{-1}$. Denote $\widetilde{s} = \bigl(D_x\phi_1 \bigr)^{-1} \s $.  $u^i$ is the $i$th element of vector
$$
u =-\sqrt{\det{\bigl(D_x\phi_1 \bigr)^{-1}} }(D_x \phi_1)^{-\top} \nabla_{\widetilde{s}} \left[ \frac{\bpsi \bigl(\frac{\widetilde{s}}{\|\widetilde{s}\|}, x \bigr)}{\sqrt{\| \widetilde{s} \|^{3}}} \right] \ .
$$

In sum, $\nabla_{\phi_1}E$ can be computed by integrating $\nabla_{\phi_1} E_x$ over the image space and written as 
\begin{align}
\label{eqn:gradE}
&\nabla_{\phi_1}E = 2 \int_{x \in \bOmega} \det(\phi_1(x)) \left\{(D_{x}\phi_1)^{-\top}
\vphantom{\left[\begin{array}{c} \divo\bigl(\bigl\langle \log_{A\bpsi_{\temp}(x)} \bpsi_{\targ}(\phi_1(x)), L_x^1 \bigr\rangle_{ A\bpsi_{\temp}(x)}\bigr) \\ \divo\bigl(\bigl\langle \log_{A\bpsi_{\temp}(x)} \bpsi_{\targ}(\phi_1(x)), L_x^2 \bigr\rangle_{ A\bpsi_{\temp}(x)}\bigr)  \\ \divo\bigl(\bigl\langle \log_{A\bpsi_{\temp}(x)} \bpsi_{\targ}(\phi_1(x)), L_x^3 \bigr\rangle_{ A\bpsi_{\temp}(x)}\bigr) \end{array}\right]}\right.\\
&\left[\begin{array}{c}\langle\;\log_{A\bpsi_{\temp}(x)}\bpsi_{\targ}(\phi_1(x))\;,\;  \frac{1}{|\triangle e_1|}\log_{A\bpsi_{\temp}(x)}A\bpsi_{\temp}(x+\triangle e_1) \;\rangle_{A\bpsi_{\temp}(x)}\\ \nonumber
\langle\;\log_{A\bpsi_{\temp}(x)}\bpsi_{\targ}(\phi_1(x))\;,\;  \frac{1}{|\triangle e_2|}\log_{A\bpsi_{\temp}(x)}A\bpsi_{\temp}(x+\triangle e_2) \;\rangle_{A\bpsi_{\temp}(x)}\\ 
\langle\;\log_{A\bpsi_{\temp}(x)}\bpsi_{\targ}(\phi_1(x))\;,\;  \frac{1}{|\triangle e_3|}\log_{A\bpsi_{\temp}(x)}A\bpsi_{\temp}(x+\triangle e_3) \;\rangle_{A\bpsi_{\temp}(x)}\end{array}\right] \\ \nonumber
& + \left.\left[\begin{array}{c} \divo\bigl(\bigl\langle \log_{A\bpsi_{\temp}(x)} \bpsi_{\targ}(\phi_1(x)), L_x^1 \bigr\rangle_{ A\bpsi_{\temp}(x)}\bigr) \\ \divo\bigl(\bigl\langle \log_{A\bpsi_{\temp}(x)} \bpsi_{\targ}(\phi_1(x)), L_x^2 \bigr\rangle_{ A\bpsi_{\temp}(x)}\bigr)  \\ \divo\bigl(\bigl\langle \log_{A\bpsi_{\temp}(x)} \bpsi_{\targ}(\phi_1(x)), L_x^3 \bigr\rangle_{ A\bpsi_{\temp}(x)}\bigr) \end{array}\right] \right\}dx \ . \nonumber
\end{align}

\begin{figure}[htb]
\raisebox{0in}[8in]
{\begin{minipage}[t]{\linewidth}
\centering
\includegraphics[height=\textheight]{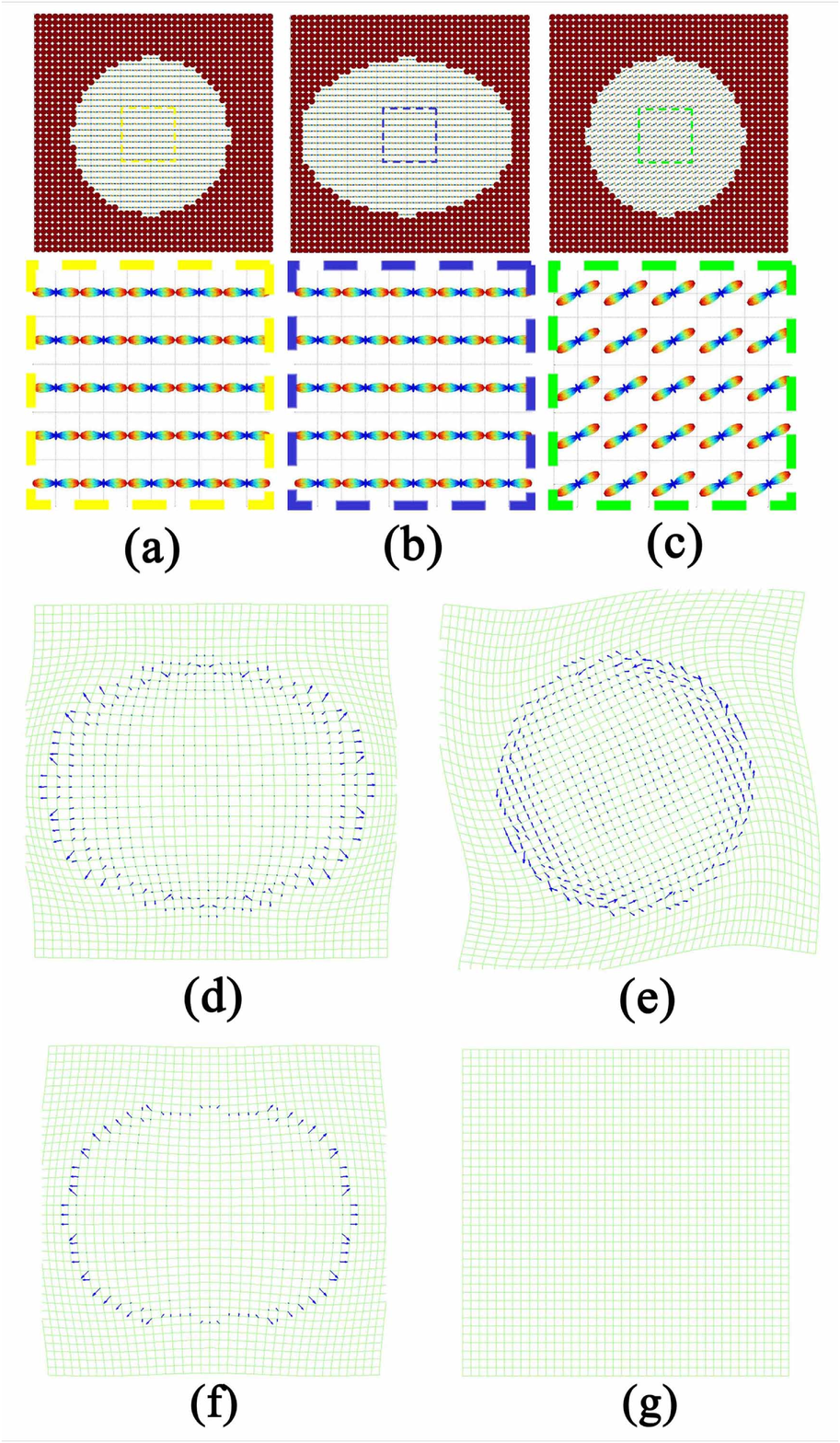}
\caption{The first and second rows respectively illustrate the original HARDI and their enlarged images. Compared to the image on panel (a), the image on panel (b) has the same ODFs but a different ellipsoidal image shape, while the image on panel (c) shows different ODFs but the same circular image shape. Panels (d) and (e) show the deformations and the corresponding momenta, calculated using $\nabla_{\phi_1}E$ in Eq. \eqref{eqn:gradE}, for mapping the image on panel (a) to panels (b) and (c), respectively.
Panels (f) and (g) show the deformations and the corresponding momenta, calculated using the gradient in our previous work \cite{Du:HARDI}, for mapping the image on panel (a) to panels (b) and (c), respectively.
}\label{fig:grad}
\end{minipage}}
\end{figure}

We now like to emphasize the difference of this above gradient derivation from our previous work \cite{Du:HARDI}. 
The fundamental difference is that in \cite{Du:HARDI}, we assume that $A$ does not change under the variation $\phi_1^\epsilon$ and thus, do not consider the variation in $A$, \ie $A^\epsilon$ is ignored. Therefore, in \cite{Du:HARDI}, the gradient of $E$ with respect to $\phi_1$ only incorporates term (A) of Eq. \eqref{eq:gradEx}. This term is similar to the scalar image matching case and only takes into account image shape difference in the volume space. We illustrate this n Figure \ref{fig:grad}, where we have one template image and two target images.
Figure \ref{fig:grad} (a) shows the template image, where its overall image shape is circular and the ODFs at each voxel inside the circle are oriented horizontally.
Figure \ref{fig:grad} (b) shows the first target image, where its overall image shape is an ellipsoid and the ODFs inside its voxels are oriented horizontally.
Figure \ref{fig:grad} (c) shows the second target image, where its overall image shape is circular as the template image but the ODFs at each voxel inside the circle are oriented at $45^\circ$. The results obtained using only term (A) as proposed in \cite{Du:HARDI} are shown in Figures \ref{fig:grad} (f, g). In Figure \ref{fig:grad} (f), we see that because of the contribution of term (A) in Eq. \eqref{eq:gradEx}, the deformation field and its corresponding momentum in the target space point to the direction that enlarges the circle to the ellipsoid.  However, in Figure \ref{fig:grad} (g), we see that term (A) in Eq. \eqref{eq:gradEx} is unable to account for such deformations as the image shapes are the same, resulting in the deformation field being zero. 
Figures \ref{fig:grad} (d, e) show the results using both terms (A) and (B) as proposed in this current paper. From Figure \ref{fig:grad} (d), we see that 
the proposed algorithm gives a deformation field that enlarges the circle to the ellipsoid, similar to that of Figure \ref{fig:grad} (f). More importantly, as shown in Figure \ref{fig:grad} (e), we see that the deformation 
that amounts to rotating the ODFs is captured by term (B) of Eq. \eqref{eq:gradEx}, which is a property that \cite{Du:HARDI} does not possess.


\afterpage{\clearpage}

\subsection{Numerical Implementation}
We so far derive $J$ and its gradient $\nabla J(m_t)$ in the continuous setting. In this section, we elaborate the numerical implementation of our algorithm under the discrete setting, in particular, the numerical computation of $\nabla_{\phi_1} E$. 

In discretization of the spatial domain, we first represent the ambient space, 
$\Omega$,
using a finite number of points on the image grid, $\Omega \cong \{(x_i)_{i=1}^N\}$. In this setting, we can assume $m_t$ to be the sum of Dirac measures such that $m_t= \sum_{i=1}^N \alpha_i(t) \otimes \delta_{\phi_t(x_i)}$ such that 
\begin{align*}
\rho(\bpsi_{\temp},\bpsi_{\targ})^2= \int_0^1 \sum_{i=1}^n \sum_{j=1}^n \alpha_i(t)^\top\bigl[k_V\bigl(\phi_t(x_i),\phi_t(x_j)\bigr)\alpha_j(t)\bigr],
\end{align*}
where $\alpha_i(t)$ is the  momentum vector at $x_i$ and time $t$.
In discretization of the spherical domain $\Do^2$, we discretize it into $N_S$ equally distributed gradient directions on the sphere. For each gradient direction $k$, it can be represented as $3$D vector with unit length $\s_k$ in Cartesian coordinate and $(r_k,\theta_k,\varphi_k)$ in the spherical coordinate. We use a conjugate gradient routine to perform the minimization of $J$ with respect to $\alpha_i(t)$. We summarize steps required in each iteration during the minimization process below: 
\begin{enumerate}
\item Use the forward Euler method to compute the trajectory based on the flow equation
\begin{align}
\label{eqn:flowEquation}
\frac{d\phi_t(x_i)}{dt} = \sum_{j=1}^N k_V(\phi_t(x_i),\phi_t(x_j)) \alpha_j(t) \ .
\end{align}

\item Compute $\nabla_{\phi_1(x_i)} E$ in Eq. \eqref{eqn:gradE}, which is described in details below.
\item Solve $\eta_t =[\eta_i(t)]_{i=1}^N$ in Eq. \eqref{eqn:eta} using the backward Euler integration, where $i$ indices $x_i$.
\item Compute the gradient $\nabla J(\alpha_i(t)) = 2 \alpha_i(t) + \eta_i(t)$.
\item Evaluate $J$ when $\alpha_i(t)= \alpha^{\text{old}}_i(t)
- \epsilon \nabla J(\alpha_i(t)) $, where $\epsilon$ is the adaptive step size determined by a golden section search. 
\end{enumerate}
Since steps $1, 3-5$ only involve the spatial information, we follow 
the numerical computation proposed in the previous LDDMM algorithm \cite{anqi_joan_2007}.  
 
We now discuss how to compute $\nabla_{\phi_1(x_i)} E$ in Eq. \eqref{eqn:gradE}, which involves the $\gsODF$ interpolation in the spherical coordinate for $A\bpsi_{\temp}(x_i)$ at a fixed $x_i$ and the $\gsODF$ interpolation in the image spatial domain for $\bpsi_{\targ}(\phi_1(x))$. To do so, we rewrite $A\bpsi_{\temp}(x_i)$ as $A\bpsi_{\temp}(\s_k, x_i)$ and $\bpsi_{\targ}(\phi_1(x_i))$ as $\bpsi_{\targ}(\s_k, \phi_1(x_i))$. For the $\gsODF$ interpolation in the spherical coordinate for $A\bpsi_{\temp}(x_i)$ at a fixed $x_i$, we  compute $A\bpsi_{\temp}(\s_k, x_i)$ according to Eq. \eqref{eq:theorem} using angular interpolation on $\Do^2$ based on spherical harmonics. 
For the $\gsODF$ interpolation in the image spatial domain for $\bpsi_{\targ}(\phi_1(x))$, we compute $\bpsi_{\targ}(\s_k, \phi_1(x_i))$ under the Riemannian framework  in \S \ref{sec:RieODF} as
\begin{align*}
\bpsi_{\targ}\bigl(\s_k,\phi_1(x_i)\bigr)=\exp_{\bpsi_{\targ}\bigl(\s_k, \phi_1(x_i)\bigr)}\sum_{j\in {\mathcal{N}}_i}w_j\log_{\bpsi_{\targ}\bigl(\s_k,\phi_1(x_i)\bigr)}(\bpsi_{\targ}(\s_k,x_j)\bigr) \ ,
\end{align*}
where ${\mathcal{N}}_i$ is the neighborhood of $x_i$, and $w_j$ is the weight of $x_j$ based on the distance between $\phi_1(x_i)$ and $x_j$. The exponential maps and logarithm maps can be computed via Eq. \eqref{eq:exp} and Eq. \eqref{eq:log} respectively. 
Finally, the inner product in Eq. \eqref{eqn:gradE}, 
\begin{align*}
\langle\;\log_{A\bpsi_{\temp}(x)}\bpsi_{\targ}(\phi_1(x))\;,\;  \frac{1}{|\triangle e_i|}\log_{A\bpsi_{
\temp}(x)}A\bpsi_{\temp}(x+\triangle e_i) \;\rangle_{A\bpsi_{\temp}(x)},
\end{align*}
and 
\begin{align*}
\langle\;\log_{A\bpsi_{\temp}(x)} \bpsi_{\targ}(\phi_1(x)) \;,\; L_x^i \bigr\rangle_{ A\bpsi_{\temp}(x)}
\end{align*}
can be computed using Eq. \eqref{eq:FisherRao}, where $\triangle e_i$ is the voxel size.

\section{Results}
\label{sec:expts}
In this section, we illustrate how LDDMM-ODF performs on both synthetic and children brain HARDI data and then compare its performance over registration based on using diffusion tensors or fractional anisotropic (FA) scalar images.

\subsection{Synthetic Data}
\begin{figure}[htb]
\centering
\includegraphics[width=0.9\linewidth]{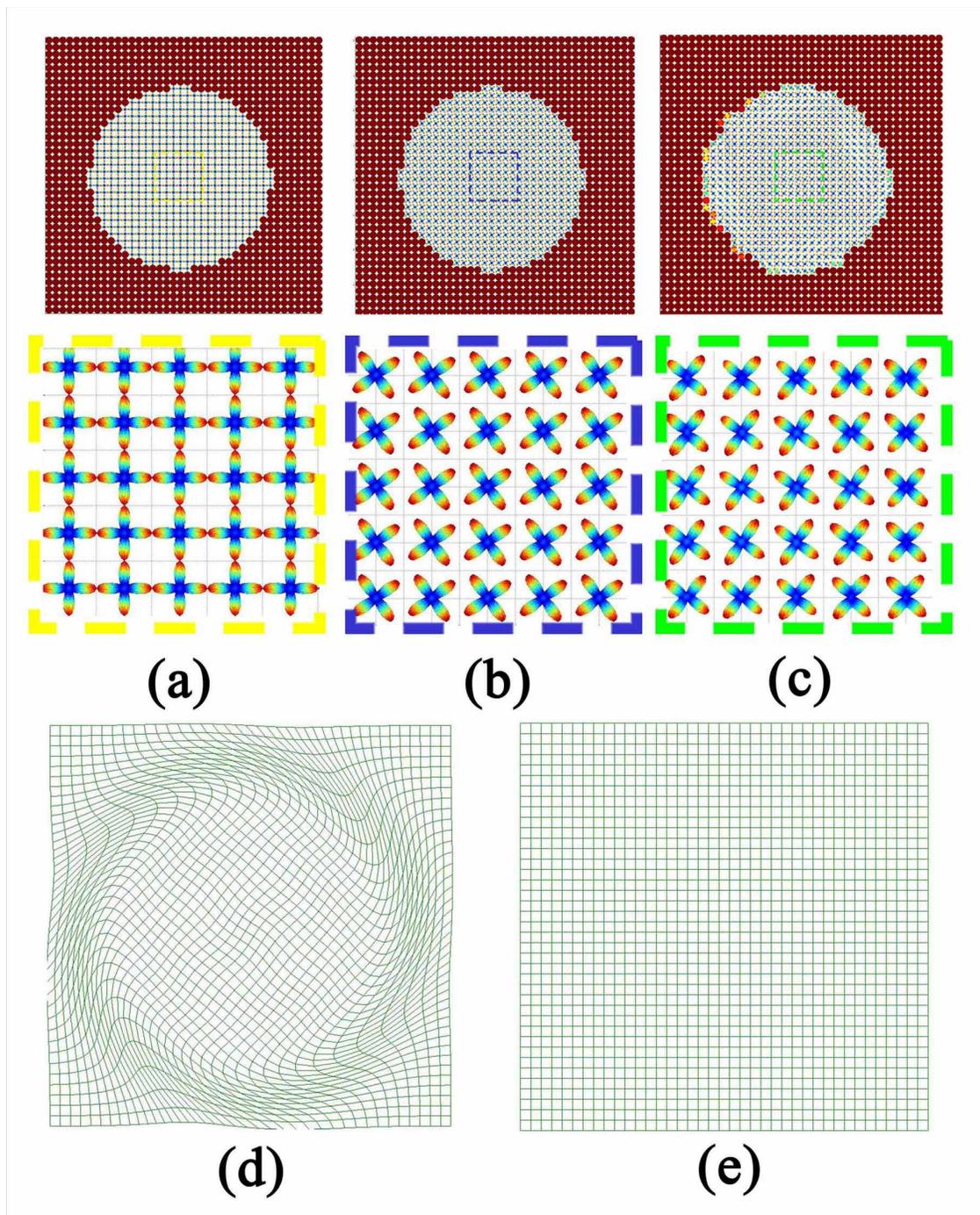}
\caption{Comparison between the LDDMM-ODF and LDDMM-DTI algorithms. Panels (a, b) respectively show the template and target HARDI and their enlarged images, where the ODF or diffusion tensor at each location contains two crossing fibers with equal orientation distribution. Panel (c) illustrates the template HARDI image transformed via the deformation given in panel (d), the result of the LDDMM-ODF algorithm. Panel (e) illustrates no deformation found via the LDDMM-DTI algorithm and thus the template HARDI image remains.
}\label{Fig:DTIvsODF}
\end{figure}

We first illustrate that the HARDI model is useful to align crossing fibers, especially when crossing fibers have equal orientation distributions. To do so, we construct two synthetic datasets, template and target, where there are two identical fibers perpendicularly crossing each other (Figure \ref{Fig:DTIvsODF} (a, b)). The orientations of the two crossing fibers differ from the template image (Figure \ref{Fig:DTIvsODF} (a)) to the target image (Figure \ref{Fig:DTIvsODF} (b)). We will compare the performance of LDDMM-ODF to the LDDMM algorithm based on DTI (LDDMM-DTI) \cite{Yan:DTI}. We refer the reader to \cite{Yan:DTI} for detailed mathematical derivation for LDDMM-DTI.

In the HARDI model, such orientation differences are encoded by the ODFs, while in the DTI model, the diffusion tensors of both the template and target data look like disks, where the first two eigenvalues being equal and the third eigenvalue being almost zero. Although the overall image shapes are the same in both the template and target HARDI data, the LDDMM-ODF algorithm is able to characterize the orientation difference of the ODFs between them by generating the deformation shown in Figure \ref{Fig:DTIvsODF} (d) with the help of term (B) of Eq. \eqref{eq:gradEx}. LDDMM-DTI fails to find any deformation (Figure \ref{Fig:DTIvsODF} (e)) even though the reorientation of the tensor is taken into account in the tensor mapping. 

\afterpage{\clearpage}

\subsection{HARDI Data of Children Brains}

In this section, we apply our proposed algorithm to real HARDI data. We evaluate the mapping accuracy of our LDDMM-ODF algorithm by comparing it with the LDDMM-image mapping based on FA (LDDMM-FA) and the LDDMM-DTI mapping based on diffusion tensors using the brain datasets of $26$ young children ($6$ years old). All three algorithms are developed under the LDDMM framework as given in \S  \ref{sec:LDDMMODF} with the exception that the matching functional, $E$, is the least square difference between two image intensities for the image mapping, LDDMM-FA, and the Frobenius norm between two tensors for the DTI mapping, LDDMM-DTI. More precisely, LDDMM-FA is based on the method developed by \cite{Beg:LDDMM} and LDDMM-DTI is based on the method developed by \cite{Yan:DTI}. In our implementation however, we optimize the deformation with respect to the momentum rather than the velocity (see \cite{Du:SD}). 
It is important to note that all three mapping algorithms used in the following evaluation have the same numerical scheme, such that any potential errors due to numerical related issues are avoided and we can make a fair comparison. 

Our image data are acquired using a $3T$ Siemens Magnetom Trio Tim scanner with a $32$-channel head coil at the National University of Singapore. Diffusion weighted imaging protocol is a single-shot echo-planar sequence with $55$ slices of $2.3mm$ thickness, with no inter-slice gaps, imaging matrix $96  \times 96$, field of view $220 \times 220mm^2$, repetition time=$6800 ms$, echo time=$89 ms$, flip angle $90^\circ$. $61$ diffusion weighted images with b=$900 s/mm^2$, $7$ baseline (b$0$) images without diffusion weighting are acquired. Notice that the b-value used in our acquisition is relatively low when compared to HARDI acquisition where $b>1000 s/mm^2$ typically. This is because the water diffusivity is in general faster in young children's brain than in adults' brain. The large b-value could result in significant loss of diffusion signals. In addition, our dataset is for the purpose of the comparison between the HARDI and DTI models. Thus, the b-value is determined by balancing the needs of both HARDI and DTI acquisition. In the data processing, DWIs of each subject are first corrected for motion and eddy current distortions using affine transformation to the b$0$ image (where there is no diffusion weighting). We randomly select one subject as the template in this study and first align the remaining subjects to this template using the affine transformation computed based on the b$0$ images of the subject and the template. Then, the DTI is computed using least square fitting \cite{Cook:Camino} and the FA is calculated from the DTI, and the ODF, $\bpsi_{\text{affine transformed}}$ , is estimated using the approach proposed in \cite{Aganj:MRM10}. We then respectively employ the LDDMM-FA, LDDMM-DTI, and LDDMM-ODF algorithms to register all subjects to the template. To ensure a fair comparison, we fix the general setting of LDDMM with kernel $\sigma_V=5$ (Eq. \eqref{eqn:lddmm}). For LDDMM-FA and LDDMM-DTI, based on the diffeomorphic mappings computed in each case, we apply the diffeomorphic group action defined in Eq. \eqref{eq:groupaction} to $\bpsi_{\text{affine transformed}}$ to obtain the registered ODFs.

\begin{figure}[htb]
\centering
\includegraphics[width=0.9\linewidth]{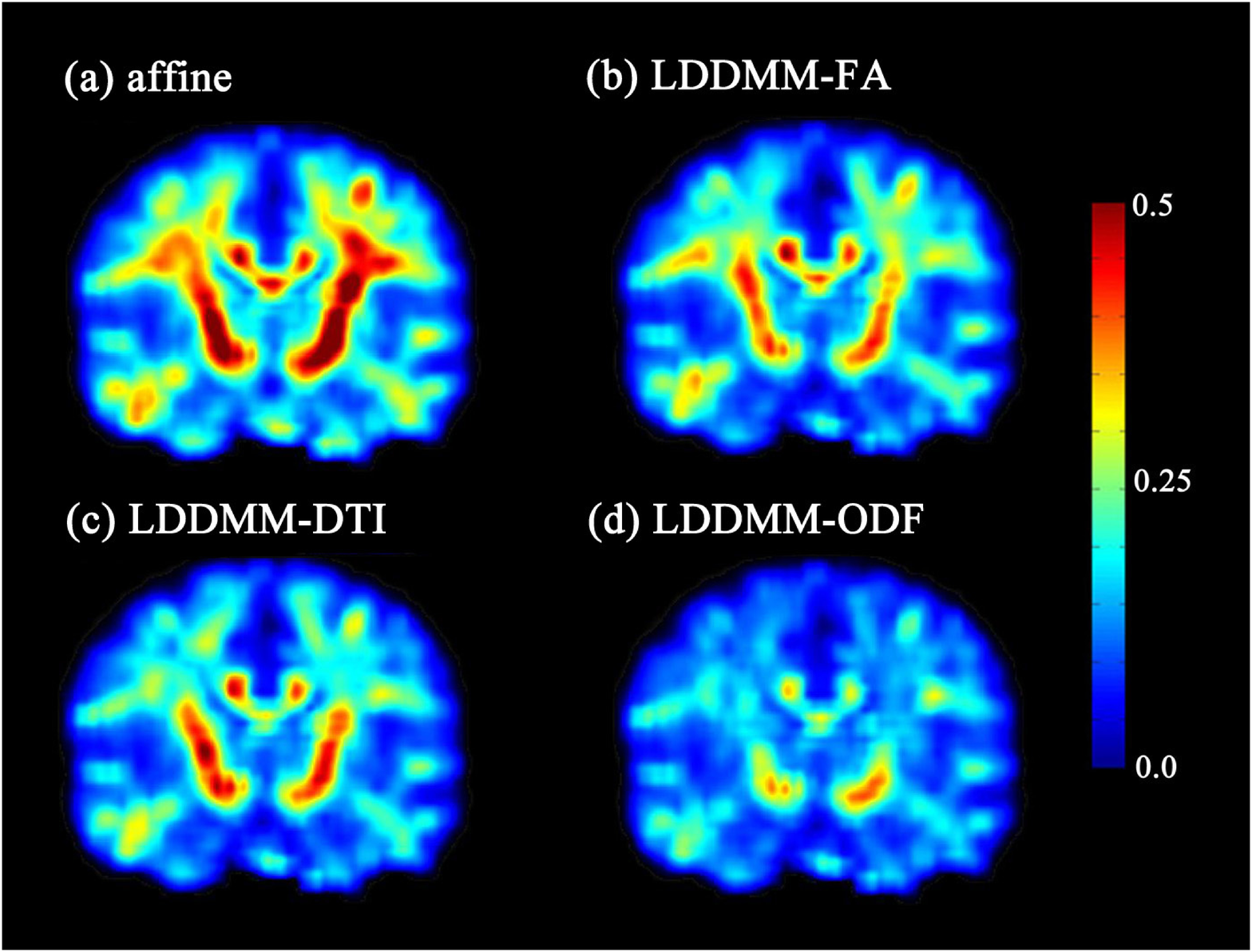}
\caption{Panels (a-d) respectively show the maps of mean symmetrized Kullback--Leibler (sKL) divergence of the ODFs between the template and the subjects deformed via affine, LDDMM-FA, LDDMM-DTI, and LDDMM-ODF.}
\label{Fig:skl_vis}
\includegraphics[width=0.7\linewidth]{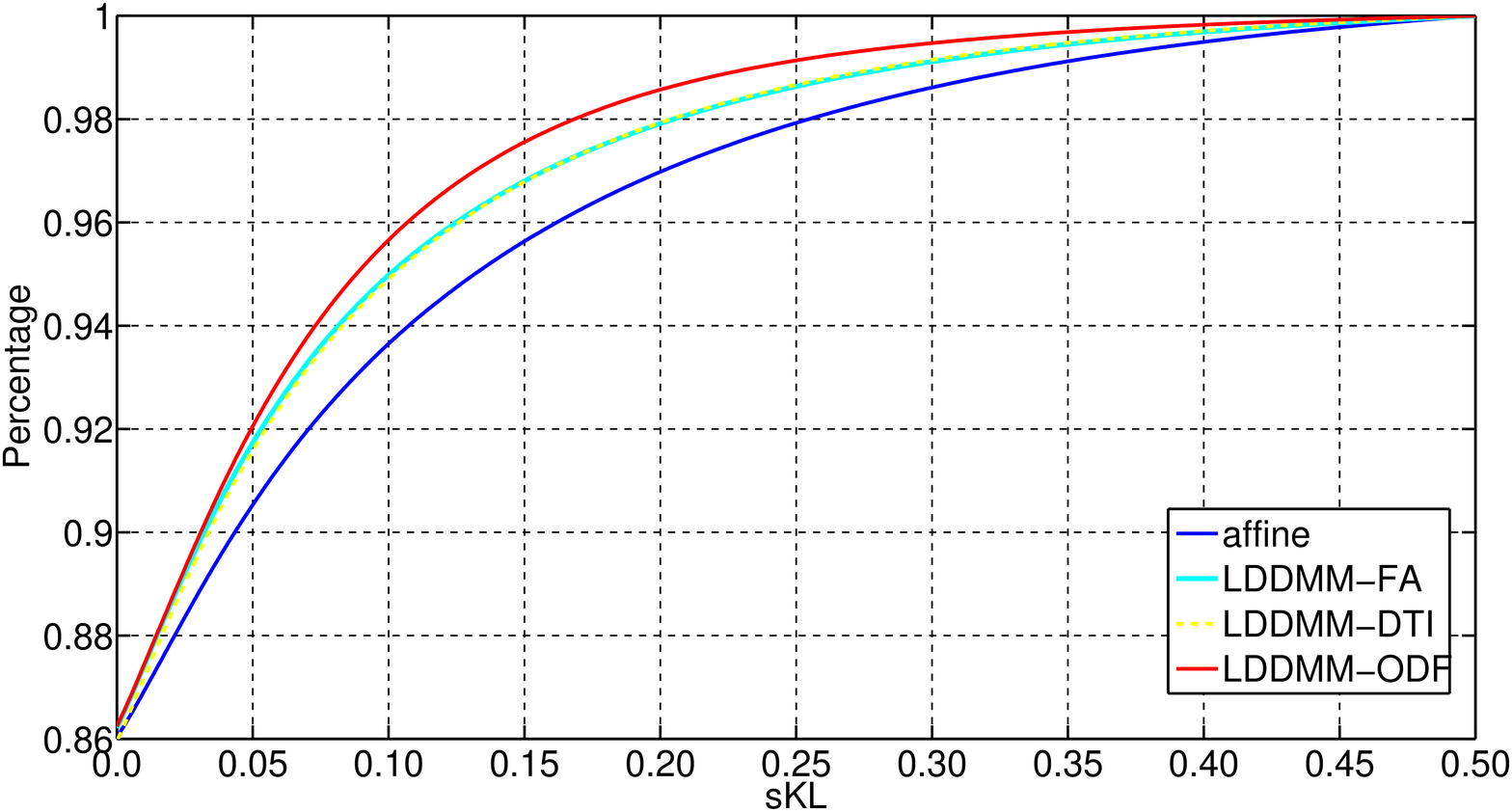}
\caption{sKL Cumulative distributions across the whole brain image and averaged over all 25 subjects are shown in blue for affine, cyan for LDDMM-FA, yellow for LDDMM-DTI, and red for LDDMM-ODF, respectively.}
\label{Fig:odf_skl_cc}
\end{figure}

To evaluate the mapping accuracy for the whole brain, we compute symmetrized Kullback-Leibler divergence (sKL) of the ODFs between the deformed subject and the template. The sKL has been used as a metric for comparing ODFs in \cite{Chiang:MICCAI08} and is defined as 
\begin{align}
\text{sKL}(\p_1,\p_2)=
\int_{\s\in\Do^2} \p_1(\s)\log \frac{\p_1(\s)}{\p_2(\s)} d\s+
\int_{\s\in\Do^2} \p_2(\s)\log \frac{\p_2(\s)}{\p_1(\s)} d\s,
\end{align}
for two ODFs $\p_1(\cdot)$ and $\p_2(\cdot)$. Lower sKL indicates that the ODF of the subjects are better aligned. Figure \ref{Fig:skl_vis} illustrates the averaged sKL maps across all $25$ subjects when affine, LDDMM-FA, LDDMM-DTI, or LDDMM-ODF are applied. This figure suggests that LDDMM-ODF is the best mapping among all studied in this paper as it has the least amount of variation, even though we do not use the sKL metric in LDDMM-ODF. Figure \ref{Fig:odf_skl_cc} also shows the cumulative distributions of sKL across the image space from each mapping. Kolmogorov-Smirnov tests on the cumulative distributions also suggest that the LDDMM-ODF significantly reduces sKL distance against the other three methods  ($p<0.001$). 

\afterpage{\clearpage}

\begin{figure}[htb]
\centering
\includegraphics[width=0.9\linewidth]{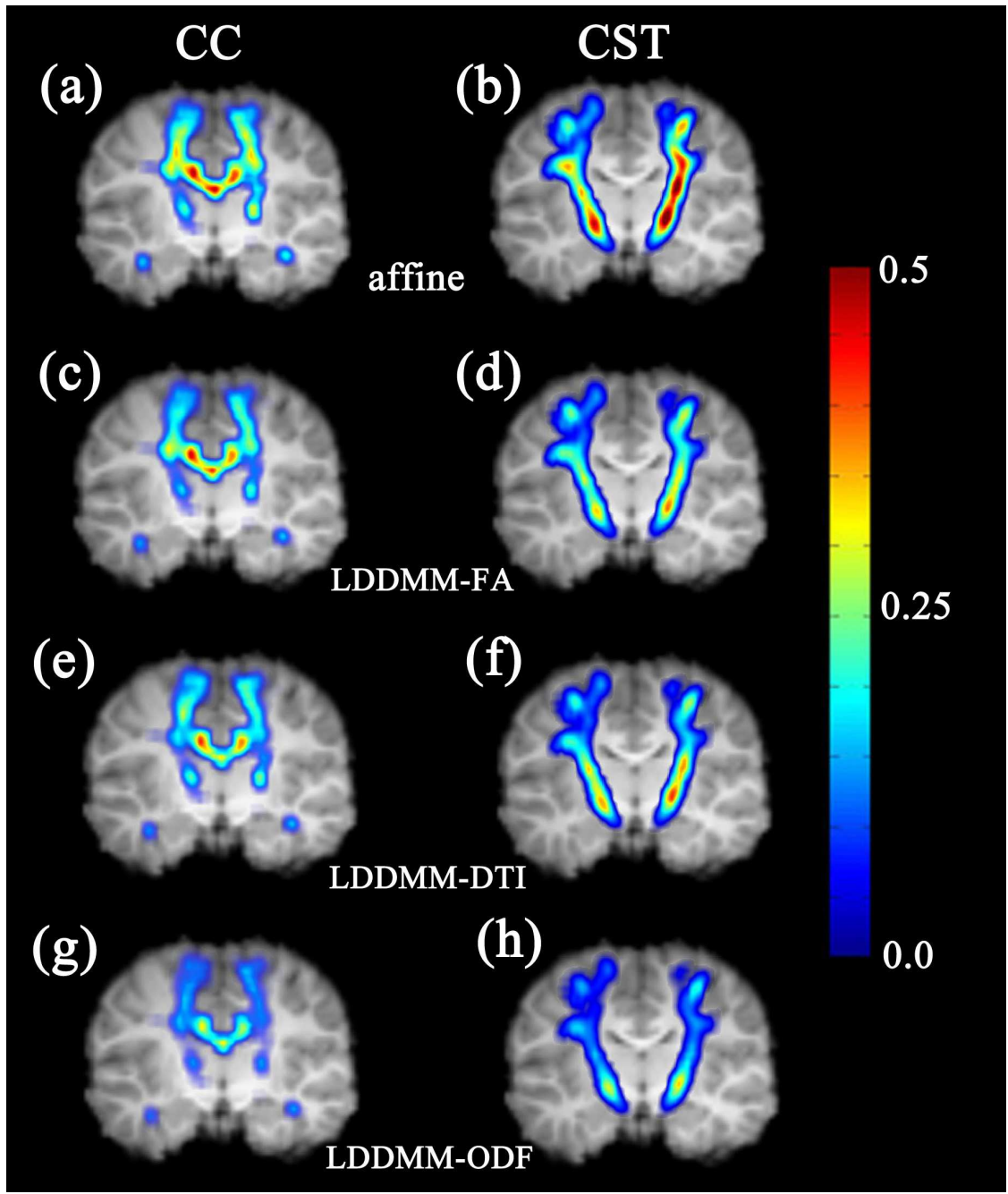}
\caption{Panels (a-h) show the maps of mean symmetrized Kullback--Leibler (sKL) divergence of the ODFs between the template and the subjects deformed via affine, LDDMM-FA, LDDMM-DTI, and LDDMM-ODF for the three major white matter tracts of the corpus callosum (CC) and bilateral corticospinal tracts (CST-left, CST-right).}
\label{Fig:tract_skl_vis}
\end{figure}

\begin{figure}[htb]
\centering
\includegraphics[width=0.9\linewidth]{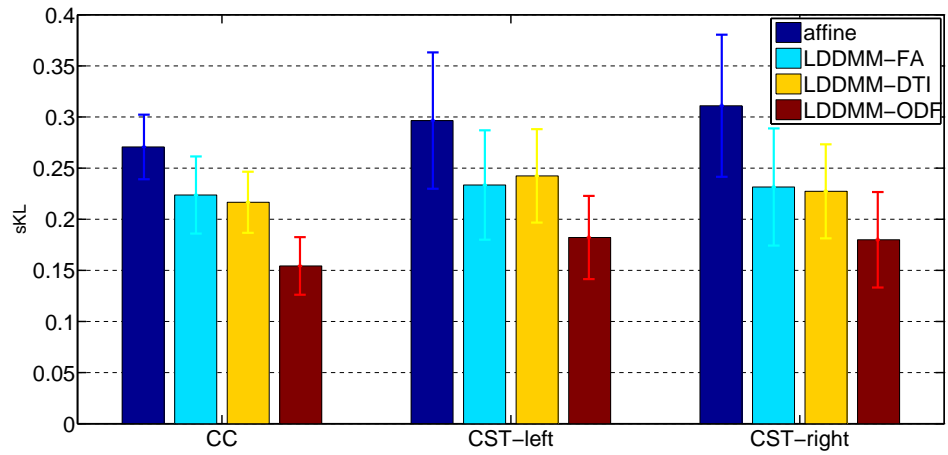}
\caption{sKL averaged over all $25$ subjects are shown for the corpus callosum (CC) and bilateral corticospinal tracts (CST-left, CST-right) when affine (blue), LDDMM-FA (cyan), LDDMM-DTI (yellow), or LDDMM-ODF (red) are applied. }
\label{Fig:tract_sKL_bar}
\end{figure}

\begin{figure}[htb]
\centering
\includegraphics[width=0.9\linewidth]{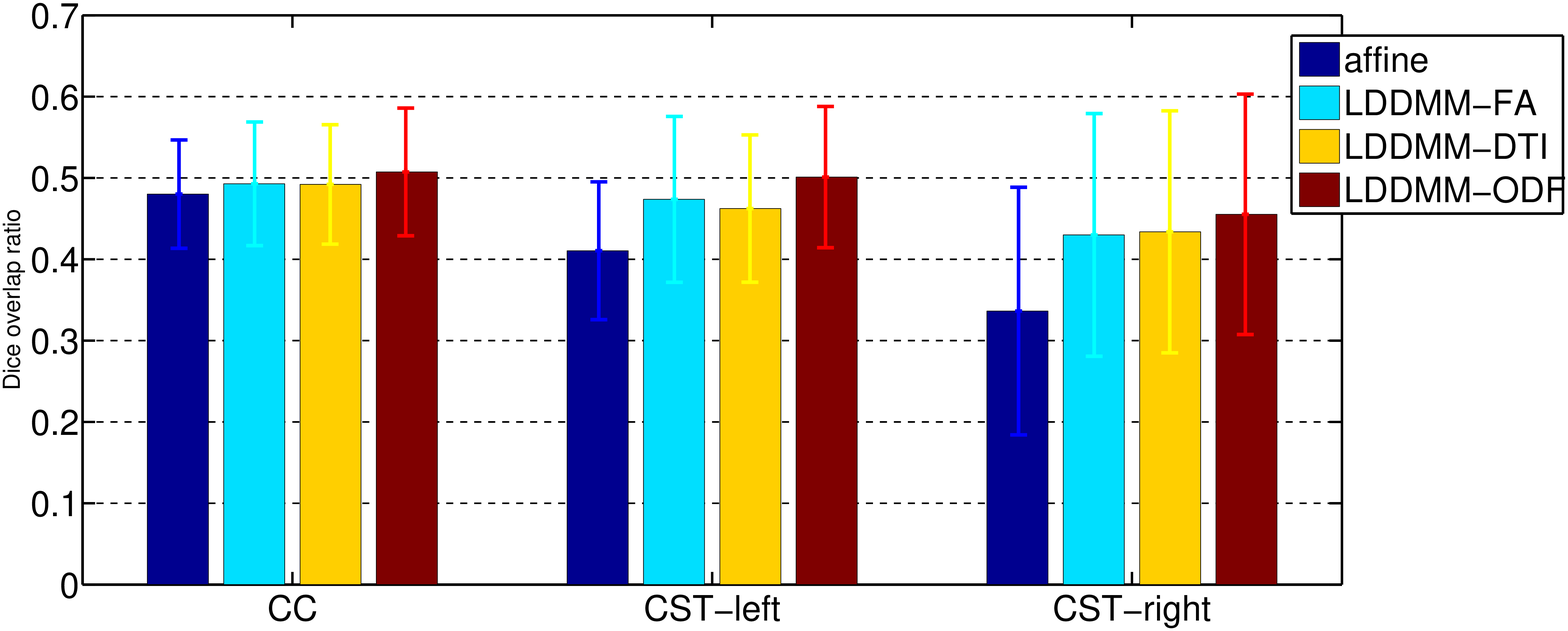}
\caption{Dive overlap ratios averaged over all 25 subjects deformed by affine (blue), LDDMM-FA (cyan), LDDMM-DTI (yellow), or LDDMM-ODF (red). }
\label{Fig:tract_dice_bar}
\end{figure}
\afterpage{\clearpage}

We now evaluate the mapping accuracy of individual white matter tracts using 1) sKL of the ODF between the template's and deformed subject's tract and 2) Dice overlap ratio to quantify the percentage of the overlap volumes between the template and deformed subject's tracts. We extract three major white matter tracts, including the corpus callosum (CC) and bilateral corticospinal tracts (CST-left, CST-right), using probabilistic tractography with the help of Camino \cite{Cook:Camino}. 
The probabilistic tractography is performed on the q-ball reconstruction using spherical harmonic representation up to order $6$ with the number of directions for each ODF limited to $3$ and the maximum allowed turning angle limited to $70^{\circ}$. 

We adopt the anatomical definition of the CC, CST-left and CST-right given in \cite{Mori:Tract} and define three regions of interest (ROI) such that each tract is comprised of all fibers passing through these three ROIs. Figure \ref{Fig:tract_skl_vis} shows the sKL maps for the three tracts, suggesting that, again LDDMM-ODF provides the best alignment for the ODFs of these three tracts when compared to affine, LDDMM-FA, and LDDMM-DTI. Figure \ref{Fig:tract_sKL_bar} shows the average sKL values for the CC, CST-left, and CST-right. Moreover, Figure \ref{Fig:tract_dice_bar} shows the averaged Dice overlap ratios across all $25$ subjects for the CC and bilateral CST. 
One-sample t-tests shows that LDDMM-ODF significantly improves the alignment of local fiber directions for three fiber tracts against the other methods in terms of sKL ($p<0.001$).
In addition, the one-sample t-tests between any two mapping algorithms suggest that all the non-linear methods show significant improvement against affine in terms of Dice overlap ratio ($p<0.001$) for the three tracts, and LDDMM-ODF shows significant improvement against LDDMM-FA and LDDMM-DTI ($p<0.001$). In the comparison between LDDMM-FA and LDDMM-DTI, the only significant difference is found in the CST-left ($p<0.05$), while no significant differences are found in the CC and CST-right.

\section{Conclusion}

We present a novel diffeomorphic metric mapping algorithm for aligning HARDI data in the setting of large deformations. Our mapping algorithm seeks an optimal diffeomorphic flow connecting one HARDI to another in a diffeomorphic metric space and locally reorients ODFs due to the diffeomorphic transformation at each location of the $3$D HARDI volume in an anatomically consistent manner. 
We incorporate the Riemannian metric for the similarity of ODFs into a variational problem defined under the LDDMM framework. The diffeomorphic metric space combined with the Riemannian metric space of ODF provides a natural framework for computing the gradient of our mapping functional. We demonstrate the performance of our algorithm on synthetic data and real brain HARDI data. 
This registration approach will facilitate atlas generation and group analysis of HARDI for a variety of clinical studies. We are currently investigating the effects of our registration algorithm on fiber tractography.


\appendices
\section{Gradient of $E_x$ with respect to $\phi_1$} 
\label{app:gradient}
We now elaborate the derivation of terms (A) and (B) in Eq. \eqref{eq:gradEx}. 

\noindent {\bf Term (A): } 
For the sake of simplicity, we denote term (A) of  Eq. \eqref{eq:gradEx} as $E_A$ and rewrite  
\begin{eqnarray*}
E_A &= &-2\Bigl\langle \log_{A\bpsi_{\temp} \circ \phi_1^{-1}(x)} \bpsi_{\targ}(x), \frac{\partial \log_{A\bpsi_{\temp} \circ \phi_1^{-1}(x)}A \bpsi_{\temp} \circ (\phi_1^\epsilon)^{-1}(x)}{\partial \epsilon}|_{\epsilon=0}  \Bigr\rangle_{A\bpsi_{\temp}\circ \phi_1^{-1}(x)} \ .
\end{eqnarray*}
Given $\frac{\partial (\phi_1+\epsilon h)^{-1}(x)}{\partial \epsilon} |_{\epsilon=0}= -\left[(D\phi_1)^{-1}h\right]\circ\phi_1^{-1}(x)$, we have 
\begin{eqnarray*}
E_A= 2\Biggl\langle \log_{A\bpsi_{\temp} \circ \phi_1^{-1}(x)} \bpsi_{\targ}(x), \Bigl\{\bigl\langle (D_x\phi_1)^{-\top}\nabla_{x}(A\bpsi_{\temp}), h\bigr\rangle\Bigr\}\circ\phi_1^{-1}(x)\Biggr\rangle_{A\bpsi_{\temp}\circ \phi_1^{-1}(x) } \ .
\end{eqnarray*}
With a change of variable from $x$ to $\phi_1^{-1}(x)$, we have 
\begin{eqnarray*}
E_A= 2 \det{\phi_1(x)}\Bigl\langle (D_x\phi_1)^{-\top} \bigl\langle \log_{A\bpsi_{\temp}(x)} \bpsi_{\targ}(\phi_1(x)), \nabla_{x}(A\bpsi_{\temp})\bigr\rangle_{ A\bpsi_{\temp}(x)}, h\Bigr\rangle \ .
\end{eqnarray*}

\noindent {\bf Term (B):}
We denote term (B) of Eq. \eqref{eq:gradEx} as $E_B$ and rewrite 
\begin{eqnarray*}
E_B & =& -2\Bigl\langle \log_{A\bpsi_{\temp} \circ \phi_1^{-1}(x)} \bpsi_{\targ}(x), 
\frac{\partial \log_{A\bpsi_{\temp} \circ \phi_1^{-1}(x)}A^\epsilon \bpsi_{\temp} \circ (\phi_1)^{-1}(x)}{\partial \epsilon}|_{\epsilon=0}  \Bigr\rangle_{A\bpsi_{\temp}\circ \phi_1^{-1}(x) }  \\
&=& -2\Bigl\langle \log_{A\bpsi_{\temp} \circ \phi_1^{-1}(x)} \bpsi_{\targ}(x), 
\frac{\partial A^\epsilon \bpsi_{\temp}(\s,x) \circ \phi_1^{-1}(x)}{\partial \epsilon}|_{\epsilon=0} \Bigr\rangle_{A\bpsi_{\temp}\circ \phi_1^{-1}(x) }  \ .
\end{eqnarray*}
According to Theorem \ref{thm:reorient}, we have 
$$
A^\epsilon \bpsi_{\temp}(\s,x) \circ \phi_1^{-1}(x)  =   \left[\sqrt{\frac{\det{\bigl(D_x\phi_1^\epsilon \bigr)^{-1}} }{\left\|{\bigl(D_x\phi_1^\epsilon \bigr)^{-1} } \s \right\|^3} } \bpsi \Bigl( \frac{(D_x\phi_1^\epsilon \bigr)^{-1} \s}{\|(D_x\phi_1^\epsilon \bigr)^{-1} \s\|}, x \Bigr)\right]\circ \phi_1^{-1}(x) \ .
$$
Denote $\widetilde{s} = \bigl(D_x\phi_1 \bigr)^{-1} \s $. Given  $\frac{\partial (D_x\phi_1^\epsilon)^{-1}(x)}{\partial \epsilon} |_{\epsilon=0}=\frac{\partial (D_x\phi_1+\epsilon D_x h)^{-1}(x)}{\partial \epsilon} |_{\epsilon=0}= -(D_x \phi_1)^{-1}D_x h(D_x \phi_1)^{-1}$, we can now compute
\begin{eqnarray*}
& &\frac{\partial A^\epsilon \bpsi_{\temp}(\s,x) \circ \phi_1^{-1}(x)}{\partial \epsilon}|_{\epsilon=0}  \\
&=&\Biggl\{ - \sqrt{\det{\bigl(D_x\phi_1 \bigr)^{-1}} }
\Bigl\langle
\nabla_{\widetilde{s}} \left[ \frac{\bpsi \bigl(\frac{\widetilde{s}}{\|\widetilde{s}\|}, x \bigr)}{\sqrt{\| \widetilde{s} \|^{3}}} \right], 
(D_x \phi_1)^{-1}D_x h(D_x \phi_1)^{-1}\s \Bigr\rangle \ \\
& & -\frac{1}{2}A\bpsi_{\temp}(\s,x) 
\tr \Bigl(D_x h (D_x \phi_1)^{-1}\Bigr)\Biggr\}\circ \phi_1^{-1}(x) \ \\
&=& 
\Biggl\{\Bigl\langle D_x h (D_x \phi_1)^{-1}\s, u \Bigr\rangle-\frac{1}{2}A\bpsi_{\temp}(\s,x)\tr \Bigl(D_x h (D_x \phi_1)^{-1} \Bigr) \Biggr\}\circ \phi_1^{-1}(x)
\ ,
\end{eqnarray*}
where  
$$ u =-\sqrt{\det{\bigl(D_x\phi_1 \bigr)^{-1}} }(D_x \phi_1)^{-\top} \nabla_{\widetilde{s}} \left[ \frac{\bpsi \bigl(\frac{\widetilde{s}}{\|\widetilde{s}\|}, x \bigr)}{\sqrt{\| \widetilde{s} \|^{3}}} \right]  \ . 
$$
We now derive the above equation in order to express it in an explicit form of $h$.  Before doing so, we first define a $3 \times 3$ identity matrix as $\Id_{3\times3}=[\e^1, \e^2, \e^3]$, where $\e^i$ is a $3 \times 1$ vector with the $i$th element as one and the rest as zero. Denote $(D_x \phi_1)^{-1}=[\w^1, \w^2, \w^3]$, where $\w^i$ is the $i$th column of $(D_x \phi_1)^{-1}$. Thus, the trace of 
$D_x h (D_x \phi_1)^{-1} $ can be written as 
$$
\tr \Bigl(D_x h (D_x \phi_1)^{-1} \Bigr) = \sum_{i=1}^3 \Bigl\langle D_xh \w^i, \e^i \Bigr\rangle \  .
$$
It yields 
\begin{eqnarray*}
& &\frac{\partial A^\epsilon \bpsi_{\temp}(\s,x) \circ \phi_1^{-1}(x)}{\partial \epsilon}|_{\epsilon=0}  \\
& = &\Biggl\{\Bigl\langle D_x h (D_x \phi_1)^{-1}\s, u \Bigr\rangle-\sum_{i=1}^3 \frac{1}{2}A\bpsi_{\temp}(\s,x) \Bigl\langle  D_xh \w^i, \e^i \Bigr\rangle  \Biggr\}\circ \phi_1^{-1}(x) \ .
\end{eqnarray*}

We introduce the following lemma \cite{Laurent_BOOK} that leads to a simple expression of $E_B$.
\begin{lemma}
For smooth vector fields, $h$, $u$, $w$, defined in a bounded open domain in  $\R^3$, 
\begin{align*}
\bigl\langle Dh\;w,u\bigr\rangle_2=-\left\langle{\left[\begin{array}{c} \divo(u^1 w) \\ \divo(u^2 w) \\ \divo(u^3 w)\end{array}\right],h}\right\rangle \ ,
\end{align*}
where $u^i$ is the $i$th element of $u$.
\end{lemma}

As a consequence, when defining $L_x^i= (D_x\phi_1)^{-1}\s u^i-\frac{1}{2}A\bpsi_{\temp}(\s,x)\w^i \ ,$ 
 it can be easily shown that 
\begin{eqnarray*}
E_B = 2\left\langle\left[\begin{array}{c} \divo\bigl( \bigl\langle \log_{A\bpsi_{\temp}(x)} \bpsi_{\targ}(\phi_1(x)), L_x^1 \bigr\rangle_{ A\bpsi_{\temp}(x)} \bigr) \\ \divo\bigl( \bigl\langle \log_{A\bpsi_{\temp}(x)} \bpsi_{\targ}(\phi_1(x)), L_x^2 \bigr\rangle_{ A\bpsi_{\temp}(x)} \bigr)  \\ \divo\bigl(\bigl\langle \log_{A\bpsi_{\temp}(x)} \bpsi_{\targ}(\phi_1(x)), L_x^3 \bigr\rangle_{ A\bpsi_{\temp}(x)} \bigr) \end{array}\right],h  \right\rangle \circ \phi_1^{-1}(x) \ .
\end{eqnarray*}
With a change of variable from $x$ to $\phi_1^{-1}(x)$, we finally have 
\begin{eqnarray*}
E_B=2\det(\phi_1(x))
\left\langle\left[\begin{array}{c} \divo\bigl( \bigl\langle \log_{A\bpsi_{\temp}(x)} \bpsi_{\targ}(\phi_1(x)), L_x^1 \bigr\rangle_{ A\bpsi_{\temp}(x)}\bigr) \\ \divo\bigl( \bigl\langle \log_{A\bpsi_{\temp}(x)} \bpsi_{\targ}(\phi_1(x)), L_x^2 \bigr\rangle_{ A\bpsi_{\temp}(x)}\bigr)  \\ \divo\bigl(\bigl\langle \log_{A\bpsi_{\temp}(x)} \bpsi_{\targ}(\phi_1(x)), L_x^3 \bigr\rangle_{ A\bpsi_{\temp}(x)}\bigr) \end{array}\right],h  \right\rangle \ .\\
\end{eqnarray*}

\section*{Acknowledgments} 
We would like to thank Alain Trouve of Ecole Normale Superieure, Cachan, France, for his very constructive and detailed comments. The work is supported by grants A*STAR SERC 082-101-0025, A*STAR SICS-09/1/1/001, the Young Investigator Award at National University of Singapore (NUSYIA FY10 P07), a center grant from the National Medical Research Council (NMRC/CG/NUHS/2010), and National University of Singapore MOE AcRF Tier 1.

\ifCLASSOPTIONcaptionsoff
  \newpage
\fi



\bibliographystyle{IEEEtran}
\bibliography{ieeetmi_hardireg}

\end{document}

%% file: HARDI_NIMG2011_macro.tex
\newcommand{\dist}{\text{dist}}
\newcommand{\bPsi}{\bm{\Psi}}

\newcommand{\bOmega}{\bm{\Omega}}

\newcommand{\bpsi}{\bm{\psi}}

\newcommand{\bxi}{\bm{\xi}}

\renewcommand{\P}{\bm{\mathcal{P}}}

\newcommand{\Do}{\mathbb{S}}
\newcommand{\p}{\bm{p}}

\renewcommand{\Re}{{\mathbb{R}}}
\newcommand{\ie}{{i.e., }}

\DeclareMathOperator*{\divo}{div}
\newcommand{\Real}{\mathbb{R}}

\newcommand{\Id}{{\mathtt{Id}}}

\newcommand{\e}{{\bf e}}
\newcommand{\R}{\Real}

\newcommand{\bL}{\mathbb{L}}

\newcommand{\temp}{\mathrm{temp}}
\newcommand{\targ}{\mathrm{targ}}
\newcommand{\tr}{\text{trace}}

\newcommand{\s}{{\bf s}}

\newcommand{\w}{{\bf w}}

\newcommand{\D}{\mathbf{D}}
\renewcommand{\v}{{\bf v}}

\newcommand{\gsODF}{\sqrt{\text{ODF}}}
\newcommand{\SPD}{\text{SPD}}

\floatstyle{ruled}
\newfloat{alg}{thp}{lop}
\floatname{alg}{Algorithm}